\documentclass[runningheads]{llncs}
\usepackage{graphicx}
\begin{document}
\title{Towards Intelligent Interactive Theatre:\\
Drama Management as a Way of Handling Performance}
\titlerunning{Towards Intelligent Interactive Theatre}
%
\author{Nic Velissaris\inst{1}\orcidID{0000-0002-5957-9892} \and \\
Jessica Rivera-Villicana\inst{1}\orcidID{0000-0003-1955-3310}}
\authorrunning{N. Velissaris and J. Rivera-Villicana}
%
\institute{Independent Researcher, Melbourne, Australia\\
\email {nic.velissaris@gmail.com, jessicarivil@gmail.com}}
\maketitle              
\begin{abstract}
In this paper, we present a new modality for intelligent interactive narratives within the theatre domain. We discuss the possibilities of using an intelligent agent that serves as a drama manager and as an actor that plays a character within the live theatre experience. We pose a set of research challenges that arise from our analysis towards the implementation of such an agent, as well as potential methodologies as a starting point to bridge the gaps between current literature and the proposed modality.

\end{abstract}
\section{Introduction}
The concept of Interactive Narrative (IN) has been pursued for several decades in different forms with the aim of providing an experience where the player feels that their decisions have an effect on the story’s development~\cite{Murray:1997:HHF:572887}. Examples of INs are the Choose Your Own Adventure (CYOA) books~\cite{Chooseco2005}, text adventures, video games like Detroit: Become Human~\cite{QuanticDream2018} and interactive films such as Black Mirror: Bandersnatch~\cite{Netflix2018}.

In this paper, we propose a novel approach to INs that consists of having an intelligent agent acting as a character in a theatre play and perform Drama Management (DM) tasks as a response to the human performers. To the best of our knowledge, such theatre modality has not yet been proposed or attempted in existing works~\cite{Pinhanez98,rousseau1997improvisational}. We propose to implement such modality using The Melete Effect by Velissaris~\cite{Velissaris2017MakingNarratives}, an IN written as a theatre play.

The novelty of the proposed approach lies in its use of principles from both “traditional” INs and theatre. While INs allow for diversity in the possible stories resulting from the actions of the user experiencing them, they require the user’s input as a participant within said story (usually as the protagonist). In theatre, the narrative typically flows linearly, with all the performers following a predefined script. As opposed to traditional INs, in our approach the user (in this case the audience) is an observer rather than a participant in the construction of the narrative. 
This involves tackling new challenges, such as the intelligent agent’s improvisation/acting skills, behaviour in accordance with character archetypes, and by the performance of DM for multiple inputs. The benefit for the audience in the creation of more sophisticated interactive experiences is a more satisfying and unpredictable narrative that can continue to change and evolve. Interactive experiences which offer genuine surprises in their outcome encourage the audience to return to the experience, leading to deeper and more satisfying engagement with the IN~\cite{szilas1999interactive}. This means that the IN has a longer `lifespan’ and can be returned to more often.

\section{Background}
Interactive Narratives: This phrase is used because it is broadly understood to mean narrative experiences that change  through the player's interaction.The mechanisms that influence interaction can vary widely from simple choices to complicated role-playing systems. Typically INs are focused exclusively on audience/player choice~\cite{SZILAS2003IDtensionDrama}. Velissaris defines this form as Choice-Based Narrative in which choice is the central mechanic that facilitates interaction~\cite{Velissaris2017MakingNarratives}. 

The Melete Effect: This IN was developed as a part of Velissaris’ doctoral research work on establishing a poetics for choice-based narratives~\cite{Velissaris2017MakingNarratives}. It tells the fictional story of a journalist Mary Melete in the 1970s and 80s. It has three distinct narratives and it will be utilised as there are no restrictions on its use.

Drama Management: From the AI perspective, a drama manager is an intelligent system which makes use of computational models of the narrative and the player in order to make choices within the environment to (attempt to) solve the boundary problem (i.e., the conflict between player agency and authorial intent) ~\cite{riedl2013interactive,sharma2010drama,Magerko:2006:PMI:1236898,weyhrauch1997guiding}.

Player Modelling: Consists of studying the interaction between a player and (usually) a game with the aim to create representations that capture desired features~\cite{Yannakakis2013,wang2017simulating}. Besides accounting for player freedom, a drama manager may be able to personalise player experience by considering the best narrative arc depending on each player’s preferences~\cite{riedl2013interactive}.

Adaptability as a Performance Trait: One of the skills an actor is taught is the ability to adapt to any situation. This adaptability is inherent to the actor’s job and any AI actor or Drama Manager will 1) need to be able to handle situations in which the experience as performed is not as was expected, and 2) adapt to these changes and ensure that experience continues on without interruption.

\section{The Challenges/Research Goals}
We now discuss the main challenges we have identified towards the realisation of this approach, and the solutions we propose.

\subsection{Exhibit believable (or up to some standard) acting skills}
Successful characterisation in a narrative experience is a result of balancing compelling actions with good performance. What a character does is measured by how it is enacted by the actor playing the role. Most of the existing work in believable characters focuses on NPCs in games~\cite{Guimaraes}. The focus on NPC believability, however, is different than ours in that believability refers to their acceptance by the player as a human-like behaving entity rather than an agent able to portray emotions or behaviour in accordance with their character archetype~\cite{Peinado2008a,PizziDavidandCharles2007}. In this regard, acting skills are related to character archetype behaviour, discussed later as a separate research goal.

The approach we propose towards achieving this objective is to have the agent learn the behaviour from human players. This could be achieved via techniques such as supervised learning~\cite{Lee2014ANarrative}. We can then apply generative models to create different courses of action for achieving similar goals that do not seem artificial (i.e, not human). The advantage of supervised learning methods is that by having a target well defined by a human, the data and training time can be reduced, while a disadvantage is the subjectivity introduced by the expert.

Another possibility is to use unsupervised learning methods, such as Apprenticeship Learning (AL) to have the agent learn a more general behavioural pattern throughout the whole story~\cite{lee2014learning}. A benefit of unsupervised learning is that there is less reliance on a human expert to dictate behaviour, but a disadvantage is that the resulting behaviour may not be of the same quality as that generated by a supervised learning method.

\subsection{Behave according to the character it’s playing}
Another challenge is to make the agent behave in character. This would require a strong modelling technique for each character tied to the computational model of the story and its constraints. We propose to build a model for each character with a representative set of traits (e.g, scales ranging from lawful to evil, specific tastes, overall role in the story, etc.) and any specific constraints regarding their behaviour (e.g, a boss that is only intolerant with their employees, when at work). This representation can be used to determine a character’s behaviour in certain scenes~\cite{PizziDavidandCharles2007,Peinado2008a}.

\subsection{Adapt to performers’ behaviour not necessarily observed in the past or planned by the author in the script}
Theatre plays are dynamic (or uncertain) to some extent. This uncertainty is expected to increase with the implementation of an IN. An AI actor will need to be able to adapt and change the narrative in ways that do not destroy the overall narrative experience. However, these changes must be in keeping with the overall narrative and be facilitated by the Drama Manager. This believable and adaptable AI actor cannot introduce plot or character details that will radically change the character in a way that threatens the coherency of the narrative.

As opposed to games, the fact that performers do not have limitations regarding the actions they can perform increases the complexity of this problem. For example, in text-based games, commands not recognised are simply not processed by the system, prompting the player to try with a different command. A factor that helps mitigate this challenge is that a scene is bounded by space and time, limiting the number of possible actions for the agent and the performers.

The approach we propose aims to have an agent whose behaviour can generalise to different situations by 1) selecting a diverse recruitment base to learn behaviour that captures a variety of possible responses to specific events, 2) implement goal/plan/action recognition to map novel events to event types that have been observed by the agent during training, and 3) encode some predefined behaviour for events that may not have been covered by the previous steps~\cite{Rivera-Villicana2018}.

\subsection{Perform drama management for more than one subject}
Existing literature focuses on managing the experience and choices of a single player. In our case, the agent needs to manage the choices of as many performers present in a scene, while the general experience is being managed for the audience.

Having multiple subjects to manage is expected to increase the complexity of the DM problem~\cite{Fairclough2003}, however, the fact that the performers possess knowledge regarding the expected outcomes of the story, and are expected to cooperate towards reaching them, may reduce the dynamism compared to traditional DM, where the player’s lack of knowledge, as well as their own preferences contribute to deviations from the author’s intended narrative. As, in our opinion, this is the most challenging goal at this time. We aim to observe Riedl et al.'s approach and evaluate its performance to find avenues for improvement~\cite{Riedl2011RobustExperiences}.

\section{Conclusion and Future Work}
The steps we propose towards achieving this proposed modality are as follows:

\begin{enumerate}
\item Using the existing script for The Melete Effect as the basis of the drama manager to develop multiple possible permutations of the narrative.
\item Learning character behaviour from actors and/or players using player modelling techniques.
\item Introducing an AI Actor in different roles to see how it responds to different story possibilities.
\item Combining the previous steps into a single system.
\end{enumerate}

In closing, we believe that this evolution of INs and AI is similar to the evolution seen in genre storytelling in other mediums. There will be many permutations and evolutionary leaps that will be required before we can establish definitively how INs and AI can work in live theatre environments. Similar to Murray’s view of the creation of Hamlet on the holodeck~\cite{Murray:1997:HHF:572887}, our aspirations are to bridge the barriers of technology and performance in a way that can revolutionise the live experience of storytelling for an audience.

%
%
%
\bibliographystyle{splncs04}
\bibliography{references.bib}

\end{document}